\pdfoutput=1
\documentclass[11pt]{article}

\usepackage[]{acl}
\usepackage{times}
\usepackage{latexsym}
\usepackage[T1]{fontenc}
\usepackage[utf8]{inputenc}
\usepackage{microtype}
\usepackage{inconsolata}
\usepackage{graphicx}
\usepackage{subcaption}
\usepackage{booktabs,longtable}

\usepackage{amsmath}
\usepackage{amssymb}
\usepackage{mathtools}
\usepackage{amsthm}
\usepackage{xspace}
\usepackage{pdfpages}
\usepackage[capitalize,noabbrev]{cleveref}

\theoremstyle{plain}

\theoremstyle{definition}

\theoremstyle{remark}

\newcommand{\system}[1]{\textsc{#1}}
\newcommand{\data}[1]{\textsc{#1}}

\newcommand{\ourdata}{\data{FabTab}\xspace}
\newcommand{\ourmethod}{\system{TAB-AUDIT}\xspace}
\newcommand{\randomforest}{\system{RandomForest}\xspace}
\usepackage{xcolor}

\definecolor{skel}{RGB}{37,92,167}
\definecolor{dig}{RGB}{178,58,46}

\title{TAB-AUDIT: Detecting AI-Fabricated Scientific Tables via Multi-View Likelihood Mismatch}

\author{Shuo Huang\footnotemark[1], 
\textbf{Yan Peng}\footnotemark[1], 
\textbf{Lizhen Qu}\footnotemark[2], 
\\
 Monash University, \\
\{shuo.huang1, lizhen.qu\}@monash.edu,
\\ypen0087@student.monash.edu }

\begin{document}

\maketitle
\footnotetext[1]{Equally Contributed.}

\footnotetext[2]{Corresponding author.}
\footnotetext[3]{The source code and dataset are available at \url{https://anonymous.4open.science/r/AI_paper_detection-3DCB/}}
\begin{abstract}
AI-generated fabricated scientific manuscripts raise growing concerns with large-scale breaches of academic integrity. In this work, we present the \textit{first} \textit{systematic} study on detecting AI-generated fabricated scientific tables in empirical NLP papers, as information in tables serve as critical evidence for claims. We construct \ourdata, the \textit{first} benchmark dataset of fabricated manuscripts with tables, comprising 1,173 AI-generated papers and 1,215 human-authored ones in empirical NLP. Through a comprehensive analysis, we identify systematic differences between fabricated and real tables and operationalize them into a set of discriminative features within the \ourmethod framework. The key feature, \textit{within-table mismatch}, captures the perplexity gap between a table’s skeleton and its numerical content. Experimental results show that RandomForest built on these features significantly outperform prior state-of-the-art methods, achieving 0.987 AUROC in-domain and 0.883 AUROC out-of-domain. Our findings highlight experimental tables as a critical forensic signal for detecting AI-generated scientific fraud and provide a new benchmark for future research\footnotemark[3].
\end{abstract}

\section{Introduction}

Large language models (LLMs) have introduced new avenues for breaching academic integrity by enabling the rapid generation of convincing but \textit{fabricated} scientific manuscripts. Qualitative studies published in \emph{Nature}~\citep{gibney2026fictionalpaper} and ~\citep{majovsky2023artificial} have shown that all major LLMs can potentially assist in producing fraudulent papers \textit{at scale}, and the resulting outputs can be sufficiently realistic to evade detection, raising serious concerns about the integrity and trustworthiness of the scientific literature~\citep{haider2024gpt,byrne2024papermills,richardson2025entities}.

Detecting AI-generated fraudulent manuscripts is increasingly challenging, as LLMs are widely used to assist with writing, clarity, and various stages of the research process~\citep{silva2025ai}. Recent advances in automated ``AI scientists''~\citep{yamada2025ai,castelvecchi2024researchers} further blur the boundary by enabling end-to-end automation of the research pipeline, from literature review to experimentation and manuscript drafting. Consequently, distinguishing fabricated AI-generated content from legitimate AI-assisted or AI-produced research becomes difficult. Inspired by prior work on detecting ghost citations~\citep{xu2026ghostcite}, we focus on \textit{detecting AI-generated fabricated tables} as a key signal of fabricated papers, since experimental tables provide transparent empirical evidence for claims and enable comparison across methods; fabricating such information fundamentally undermines the credibility of the entire manuscript.

Despite growing concerns about AI-generated fraudulent content in scientific research, there has been no systematic study on evaluating and detecting AI-generated fabricated experimental tables. Prior work has primarily focused on textual artifacts, such as hallucinated citations or generated narratives~\citep{gao2023comparing,haider2024gpt}, while largely overlooking structured empirical evidence in tables. Furthermore, the lack of quantitative analyses and benchmark datasets limits our ability to characterize the prevalence, patterns, and detectability of fabricated tables, thereby hindering the development of reliable detection methods.

\begin{figure}[t]
\centering
\setlength{\tabcolsep}{4.1pt}
\renewcommand{\arraystretch}{1.04}
{\footnotesize
\begin{tabular}{@{}lccc@{}}
\toprule
\textcolor{skel}{\textbf{Model}} & \textcolor{skel}{\textbf{P}} & \textcolor{skel}{\textbf{R}} & \textcolor{skel}{\textbf{F1}}\\
\midrule
\textcolor{skel}{BERT-base} & \textcolor{dig}{72.4} & \textcolor{dig}{70.8} & \textcolor{dig}{71.6}\\
\textcolor{skel}{SpanBERT} & \textcolor{dig}{76.9} & \textcolor{dig}{75.8} & \textcolor{dig}{76.3}\\
\textcolor{skel}{Ours} & \textcolor{dig}{\textbf{98.7}} & \textcolor{dig}{\textbf{98.4}} & \textcolor{dig}{\textbf{98.6}}\\
\bottomrule
\end{tabular}
}

\vspace{0.18em}
{\scriptsize
\textcolor{skel}{Blue} = skeleton (headers/row labels)\qquad
\textcolor{dig}{Red} = numeric tokens
}

\vspace{0.06em}
{\scriptsize
\textbf{Mismatch} $\uparrow$: typical \textcolor{skel}{skeleton}, atypical \textcolor{dig}{98.x} digits.
}

\caption{Compact illustration of the table skeleton--numeric mismatch.}
\label{fig:intro-mismatch-compact-demo}
\end{figure}

To enable systematic investigation and evaluation of AI-fabricated tables, we construct the \textit{first} benchmark dataset of \underline{fab}ricated manuscripts with \underline{tab}les, named \ourdata, comprising 1,173 AI-generated papers and 1,215 human-authored ones in empirical NLP. We focus on this paper type because tables are the primary medium for presenting evidence supporting key claims. We then conduct a comprehensive analysis of the differences between fabricated and real tables and operationalize those findings into a set of highly discriminative features in the framework \ourmethod. We explore the inherent inconsistency between table skeleton and numeric digits of the table as shown in Figure.~\ref{fig:intro-mismatch-compact-demo}. Based on these features, we show that traditional ML models significantly outperform prior SOTA methods~\citep{hans2024binoculars,bao2024fastdetectgpt} by a wide margin.


Our key contributions are three-fold:
\begin{itemize}
    \item We construct \ourdata using a dedicated pipeline that leverages literature on the same topics. The in-domain set comprises 1,012 recent human-authored papers and 976 fabricated ones generated by \system{GPT-4o} conditioned on existing literature. To reflect real-world difficulty, each generated paper is paired with real papers on closely related topics. In total, the in-domain set contains 5,284 tables. To evaluate generalization to unseen generation models, we build an out-of-domain test set using \system{GPT-5.2}, consisting of 197 fabricated papers and 203 human-authored ones on highly similar topics. A human study (\S\ref{sec:human_study}) further shows that papers generated by our pipeline are difficult to distinguish from real ones.
    
    \item Based on a comprehensive analysis of the dataset, we operationalize our findings into a set of highly discriminative features within the \ourmethod framework. The most salient feature is \textit{within-table mismatch}, which captures the perplexity gap between a table's structural skeleton and its numerical content. We also incorporate additional numerical features that do not rely on language models, which further enhance detection performance. 
    \item We train traditional machine learning models on our features aggregated at the paper level and evaluate them on both in-domain and out-of-domain test sets. Among all models, \randomforest~\citep{rigatti2017randomforest} achieves the best performance, with 0.987 AUROC on the in-domain test set and 0.883 AUROC on the out-of-domain test set. We further show that our models achieve higher AUROC when detecting fabricated tables generated by simple prompting of LLMs compared to those produced by our paper generation pipeline.
\end{itemize}

\section{Benchmark Construction and Paper Generation}
\label{sec:generation}

Our benchmark is designed to isolate \emph{provenance} rather than topic. The human papers are composed of recent experimental NLP papers, while the fabricated arm is generated to match the same broad domain and paper format. Without it, a detector could separate classes using stale venue conventions or cross-domain artifacts instead of signals tied to fabricated empirical evidence.

\subsection{Collecting Recent Experimental NLP Papers}
We draw the human corpus from arXiv rather than from a heterogeneous mixture of venues or historical snapshots. To reduce the chance that benchmark papers were already absorbed into the pretraining data of the language models used in generation or auditing, we restrict the source pool to papers posted within the month immediately preceding collection. This recency constraint also keeps the benchmark close to a single publication regime, limiting drift in formatting conventions, writing style, and topical emphasis.

We further restrict the source pool to \emph{experimental NLP papers}. This avoids mixing result tables from substantially different scientific genres, such as purely theoretical work, surveys, or papers from unrelated domains whose tables obey different reporting conventions. In the indexed human source pool, this procedure yields 1,118 candidate papers. After table extraction and validity filtering, 1,012 human papers remain in the clean benchmark used.

This collection strategy matters for two reasons. It reduces domain bias: a detector should not win simply because fabricated papers come from a different field or historical period. Second, it raises the scientific standard of the benchmark: if a detector succeeds here, it is more likely to be detecting signals tied to fabricated empirical evidence rather than superficial topic or formatting artifacts.

\subsection{Literature-Grounded Fabricated Paper Generation}

To construct realistic yet controllable fabricated papers---primarily to obtain realistic fabricated \emph{tables}---we design a literature-grounded pipeline for empirical NLP (Appendix Figure~\ref{fig:generation_pipeline}). Recent arXiv papers are embedded and clustered by abstract, and references are sampled from a coherent research line. Their abstracts ground the generator, which is then prompted with a structured LaTeX template containing standard scientific sections and result tables. Grounding preserves topical coherence, while structural constraints improve stability for analysis. 

Although the benchmark is organized at the paper level, the realism target is the \emph{table} rather than the manuscript as a whole. We use full-paper generation as contextual scaffolding: a table embedded in a literature-grounded manuscript inherits a coherent task, baseline set, metric vocabulary, and surrounding explanation, which makes it look more scientific than an isolated table produced by direct prompting alone. This design choice improves the realism of the fabricated empirical evidence that TAB-AUDIT is meant to audit.
Accordingly, the synthetic outputs should be interpreted as \emph{literature-grounded but not experiment-grounded}. Their titles, problem settings, related-work framing, and table schemas are anchored in genuine research themes, but the reported values are synthesized rather than derived from actual experimental runs. The surrounding prose is included primarily to support realistic presentation of the tables, not to simulate a fully validated scientific contribution. An example generated paper is shown in Appendix~\ref{sec:appendix-running-examples}.
The main fabricated arm used in the labeled benchmark is generated with GPT-4o. To evaluate transfer under generator shift rather than memorization of a single generator's quirks, we additionally construct a separate GPT-5.2 fabricated holdout, which is reserved entirely for later evaluation.

\subsection{Benchmark Details}
\label{sec:benchmark_stats}

After filtering, the main labeled benchmark contains 1,012 human papers and 976 fabricated papers, for a total of 1,988 papers with 5,284 extracted tables. The external GPT-5.2 holdout adds 400 papers and 1,324 extracted tables for transfer evaluation. Furthermore, we leverage two local model QWEN2.5-7B and Llama3.1-8B as our generation models to investigate the impact of generators. As a result 351 valid AI papers are collected for QWEN2.5-7B and 237 valid AI papers with tables from Llama3.1-8B. We defer the exact train/validation/test protocol to Section~\ref{sec:experimental_setup}, but at a high level this construction gives us a recent, domain-matched, paper-level benchmark in which topic and document form are aligned across classes and the main difference of interest is empirical provenance.

\subsection{Can human really differentiate AI-fabricated tables?}
\label{sec:human_study}
To evaluate whether our AI-generated tables are distinguishable from those originating in real scientific papers by human judgment alone, we conducted a human evaluation study. The example is in Appendix.~\ref{fig:from_screenshot}. The evaluation was implemented as a quiz using Google Forms. We randomly sampled 25 tables from AI-generated scientific papers produced by our generation pipeline and 25 tables from real human-authored papers, resulting in a balanced set of 50 tables. The tables were presented in isolation and in a randomized order to avoid contextual cues or learning effects. The study involved three voluntary participants who are CS PhD candidates and was intended as a preliminary exploratory analysis.They have consented the sharing of the responses. The responses were collected in a standardized manner through the Google Forms platform.  Based on collected responses, we computed the receiver operating characteristic (ROC) curve and obtained an area under the curve (AUC) of 0.6608. Although this result should be interpreted with caution due to the small number of participants, it indicates that human judgments perform better than random guessing but achieve only moderate discriminative ability. The relatively low AUC suggests that AI-generated tables can exhibit a high degree of surface-level realism, making human discrimination challenging in the absence of additional textual or methodological context.

\section{Signal Analysis: How Fabricated Tables Differ from Human Experimental Table}
\label{sec:case_study}
The main scientific claim of this paper is stronger than "a classifier works on a benchmark." 
Our claim is that AI-fabricated tables differ from human experimental tables in a \emph{systematic and interpretable} way. 
We therefore analyze the signal before introducing the full auditing framework. 
This ordering is intentional: if the benchmark is realistic, then a useful detector should be traceable to a concrete distributional gap rather than to opaque classifier behavior.

\subsection{Why a forensic signal should exist}

A scientific result table is not just a collection of plausible numbers. 
It is the visible trace of an underlying measurement process: model training, evaluation, comparison against baselines, and decisions about what to report. 
As a result, genuine tables couple two channels that are produced together: a \emph{textual skeleton} (titles, headers, row labels, task names, metric names, and comparison structure) and a \emph{numeric channel} (the values that summarize experimental outcomes).

This coupling is difficult to reproduce perfectly through generation alone. 
An LLM can imitate the textual skeleton of an NLP table by borrowing familiar benchmark names, model families, and formatting conventions from the literature. 
However, matching the \emph{numeric texture} of genuine experimental reporting is harder. 
Real tables often reflect uneven gains, local trade-offs, inconsistent decimal precision, and incidental arithmetic relations that arise from actual measurement rather than from direct synthesis. 
This suggests a forensic signal: fabricated tables may retain a highly conventional scientific skeleton while their numbers remain atypical \emph{relative to that skeleton}.

This observation motivates a within-table comparison. 
Rather than asking whether a table looks globally probable under a language model, we ask whether its numeric channel looks typical \emph{given} the kind of textual skeleton it presents. 
That relative view is important because absolute perplexity alone is easily confounded by topic, rare benchmark names, or formatting quirks.

\subsection{Multi-view likelihood evidence for numeric--skeleton mismatch}

To probe this hypothesis, we score each extracted table under multiple observer-model views: a skeleton-only view with numbers masked, a digit-level numeric view, and a full view containing both text and numbers. 
The key quantity is the within-table mismatch
\begin{equation}
\begin{split}
f_{\Delta}(T) &= \mathrm{slog1p} \Big( \mathrm{PPL}_M(s^{\mathrm{dig}}(T)) \\
&\quad - \mathrm{PPL}_M(s^{\mathrm{text}}(T)) \Big),
\end{split}
\end{equation}
which compares how atypical the numeric channel is relative to the textual skeleton under the same observer model $M$. 
We later use this signal inside \textsc{TAB-AUDIT}; here, we examine it directly as evidence of a distributional gap.

\begin{figure}[t]
  \centering
  \includegraphics[width=0.32\linewidth]{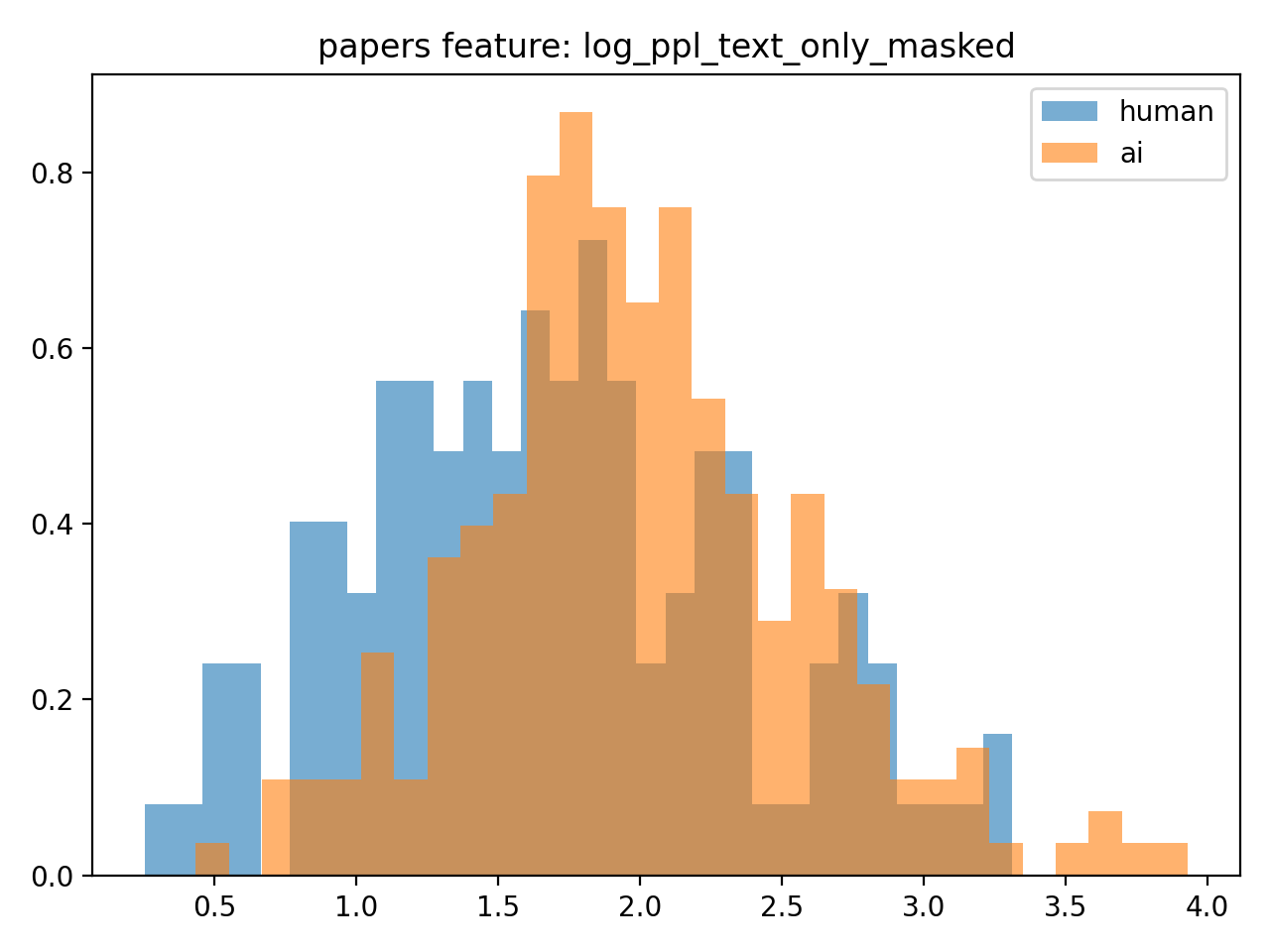}
  \includegraphics[width=0.32\linewidth]{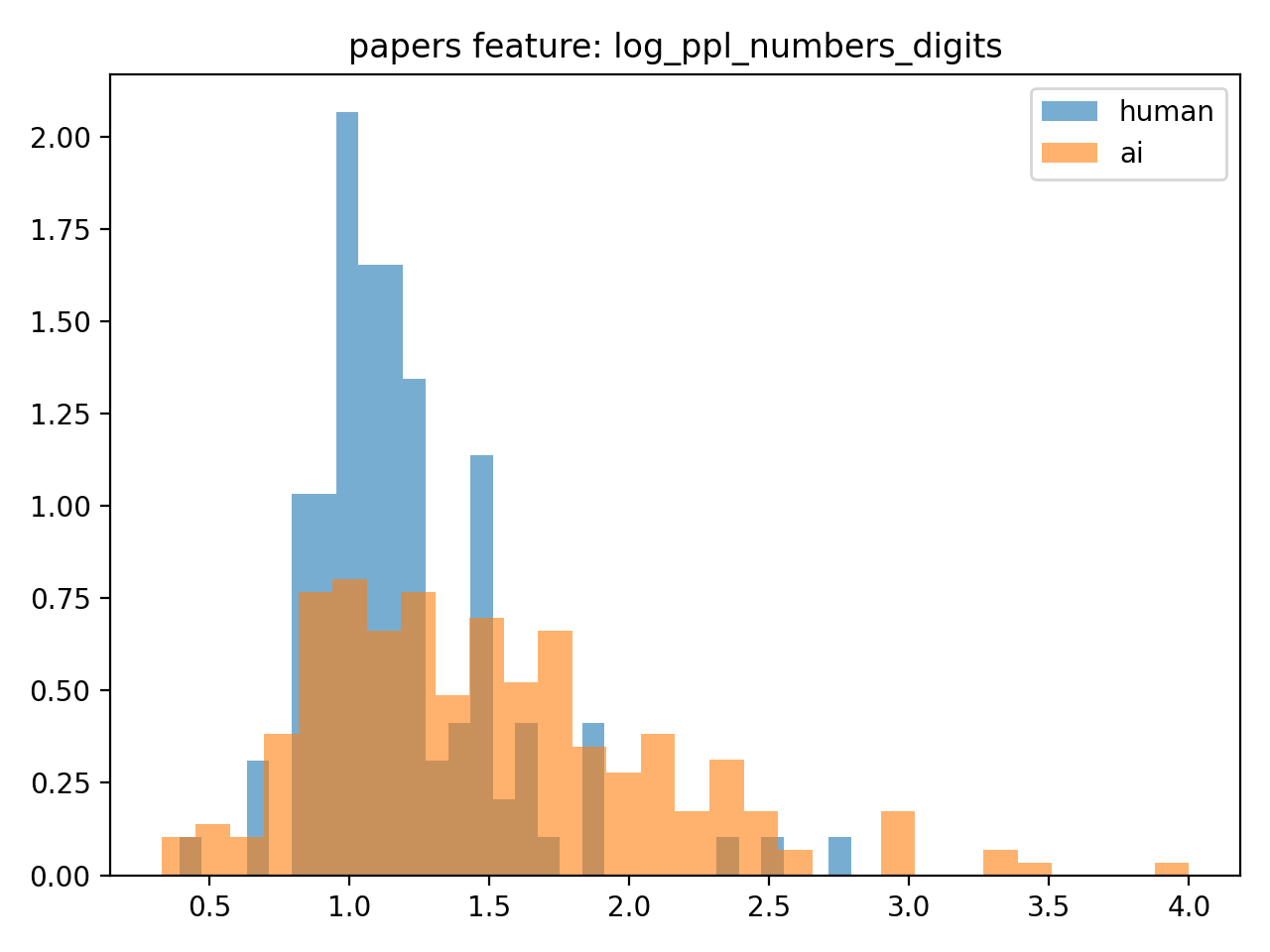}
  \includegraphics[width=0.32\linewidth]{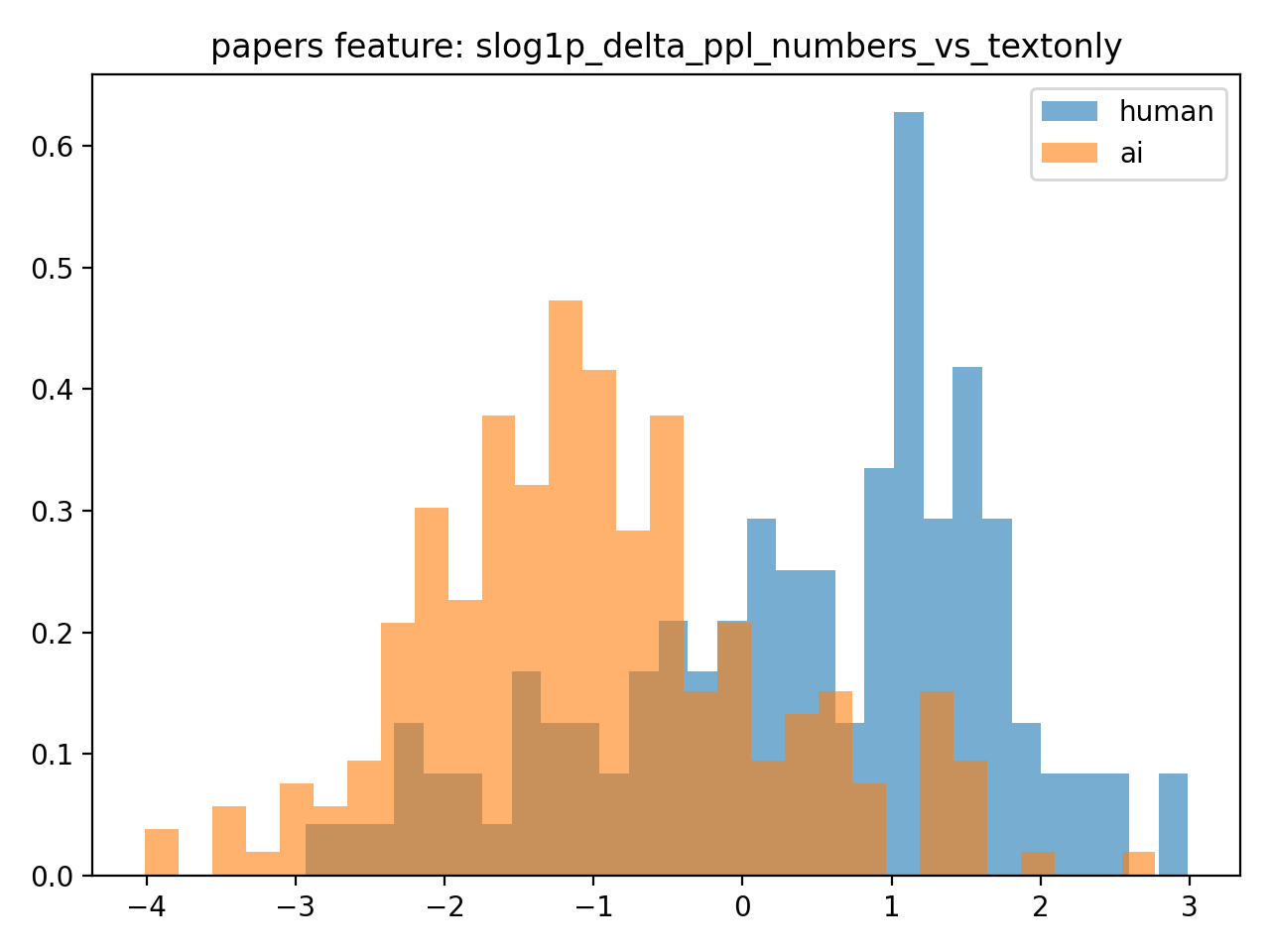}
  \caption{Representative paper-level distributions for three observer-based indicators: skeleton likelihood (left), digit-level numeric likelihood (center), and the within-table numeric--skeleton mismatch (right). The key pattern is not merely that fabricated tables can have unusual numbers in isolation, but that their numbers are atypical \emph{relative to an otherwise conventional scientific skeleton}.}
  \label{fig:case_univariate}
\end{figure}

Figure~\ref{fig:case_univariate} shows the paper-level distributions of three representative indicators. 
Two patterns are especially important. 
First, the skeleton-only likelihood exhibits substantial overlap between human and fabricated papers. 
This is itself informative: it suggests that the surface scientific skeleton is comparatively easy to imitate, so prose-like fluency alone is not a reliable authenticity cue for result tables. 
Second, the mismatch feature shows a much clearer shift. 
In fabricated papers, the numeric channel is more often atypical relative to the surrounding skeleton, whereas in human papers the two channels are better aligned. 
In other words, the tables in fabricated papers frequently look \emph{structurally right} before they look \emph{experimentally grounded}.

This analysis also clarifies why we rely on \emph{within-table} contrasts instead of raw perplexity alone. 
Absolute likelihood is affected by many nuisance factors, including table length, topic, model names, and serialization details. 
By contrasting the numeric and textual views of the \emph{same} table, we cancel much of that variation and focus the detector on provenance-relevant inconsistency. 
This is the core reason multi-view likelihood mismatch is a useful auditing signal.

\subsection{Complementary LM-free numeric signals}
\label{stats:lm-free}

We also consider statistical evidence of the tables as signals as it carries human conventions of building experimental tables.
We therefore compute a set of simple numeric diagnostics directly from table values, asking whether a table behaves like a summary of empirical comparison rather than a smooth synthetic pattern.

Table~\ref{tab:top_stats:wide} highlights three recurring regularities. 
human tables show substantially stronger \emph{arithmetic coherence}. 
For example, row $\Delta$-consistency and row \%$\Delta$-consistency are both higher for human tables, suggesting that adjacent values more often participate in meaningful differences or percentage changes induced by actual experimental comparison. 
This is what we would expect from tables that summarize ablations, improvements over baselines, or metric trade-offs.

Full analysis of this table can be found for Appendix.~\ref{app:lmfree-stats}

\subsection{Implications for auditing}

Taken together, the observer-based and LM-free analyses point to the same conclusion. 
AI-fabricated tables can imitate the \emph{form} of scientific reporting surprisingly well, but they reproduce the \emph{distributional texture} of genuine experimental evidence much less faithfully. 
This is why the problem should be framed as forensic auditing rather than generic text detection. 
The useful evidence is not only whether a table looks fluent or well formatted, but whether its numeric behavior is appropriately matched to its scientific skeleton.

\section{\textsc{TAB-AUDIT}: A Signal-Driven Paper-Level Auditing Framework}
\label{sec:method}
Section~\ref{sec:case_study} showed that fabricated tables differ from human experimental tables through a forensic signature: the textual skeleton can remain highly conventional even when the numeric channel is atypical relative to that skeleton. We now operationalize that finding in \textsc{TAB-AUDIT}. The framework is designed around \emph{signal extraction} rather than a particular classifier family. Its core output is a paper-level risk score derived from table-level provenance cues; the choice of logistic regression, Random Forest, or another detector is a secondary operational layer.

\subsection{Problem setup}

A paper $p$ contains a set of extracted tables
\[
\mathcal{T}_p=\{T_{p,1},\dots,T_{p,m_p}\}.
\]
Our task is paper-level provenance auditing: given only the published paper or its extracted tables, predict
\[
y_p \in \{0,1\},
\]
where $y_p=1$ denotes an AI-fabricated paper and $y_p=0$ denotes a human-authored empirical paper.

The auditor has no access to code, logs, checkpoints, or raw experimental outputs. This constrained setting is deliberate: the goal is not replication, but forensic screening from the published artifact alone. We assume access to an \emph{observer} language model $M$ used only to score serialized table views. We do \emph{not} treat $p_M(\cdot)$ as a calibrated generative model of scientific tables. Instead, we use the observer only to measure \emph{relative typicality} and, crucially, \emph{within-table contrasts} that are less sensitive to topic and formatting confounds than absolute likelihood alone.

\subsection{Multi-view table serialization}

Let $s(\cdot)$ denote a deterministic serialization of a table into a token sequence. For each table $T$, we construct three main views.

\textbf{Full view.}
$s^{\mathrm{full}}(T)$ linearizes the title, headers, row labels, and all cell contents. This view preserves the complete table.

\textbf{Textual-skeleton view.}
$s^{\mathrm{text}}(T)$ masks numeric substrings with a placeholder token while keeping titles, headers, row labels, metric names, and layout tokens. This isolates the scientific skeleton.

\textbf{Digit-level numeric view.}
$s^{\mathrm{dig}}(T)$ retains only the numeric substrings, rewritten at digit level (for example, \texttt{12.34} becomes \texttt{1 2 . 3 4}). This reduces tokenizer-specific artifacts and makes the numeric channel more comparable across tables.

These views are designed to separate \emph{what kind of table this is} from \emph{what numeric evidence it reports}. The first is largely answered by the textual skeleton; the second is encoded primarily in the digits.

\subsection{Likelihood-based signal extraction}

Given a serialized sequence $x=(x_1,\dots,x_n)$ and observer $M$, we compute token-level negative log-likelihood and perplexity as
\begin{align}
\mathrm{NLL}_M(x) &= -\frac{1}{n}\sum_{i=1}^{n}\log p_M(x_i \mid x_{<i}), \\
\mathrm{PPL}_M(x) &= \exp(\mathrm{NLL}_M(x)).
\label{nll}
\end{align}
For long serializations, we use a strided sliding-window implementation to avoid truncation bias, and we work with $\log \mathrm{PPL}$ where appropriate for numerical stability.

From these views, \textsc{TAB-AUDIT} forms a compact four-dimensional likelihood feature block,
\begin{equation}
\begin{split}
\phi_{\mathrm{PPL4}}(T) = \Big[ & \log \mathrm{PPL}_M(s^{\mathrm{full}}(T)), \\
                                & \log \mathrm{PPL}_M(s^{\mathrm{text}}(T)), \\
                                & \log \mathrm{PPL}_M(s^{\mathrm{dig}}(T)), f_{\Delta}(T) \Big].
\end{split}
\end{equation}
The first three terms record absolute observer typicality under the three views. The fourth term captures the core forensic signal:
\begin{equation}
\begin{split}
f_{\Delta}(T) = \mathrm{slog1p}\Big( & \mathrm{PPL}_M(s^{\mathrm{dig}}(T)) \\
& - \mathrm{PPL}_M(s^{\mathrm{text}}(T)) \Big),
\end{split}
\label{eq:fdelta}
\end{equation}
where $\mathrm{slog1p}(z)=\mathrm{sign}(z)\log(1+|z|)$.

This feature is central to the framework. A high value indicates that the numeric channel is unusually atypical relative to an otherwise conventional scientific skeleton. In our setting, this is the main provenance cue revealed by the signal analysis in Section~\ref{sec:case_study}.

\subsection{Complementary structural and LM-free numeric cues}

The likelihood mismatch signal is informative but not exhaustive. Section~\ref{sec:case_study} also showed that fabricated tables often differ from human tables in simpler, model-free ways: they can be overly regular, overly sorted, or unusually repetitive, and they may exhibit weaker arithmetic coherence. To capture these effects, \textsc{TAB-AUDIT} augments the likelihood block with two additional feature families.

\textbf{Structural features.}
We extract lightweight table-shape indicators such as row and column counts, header/body proportions, and numeric-cell density. These features capture broad reporting style without depending on lexical content.

\textbf{LM-free numeric features.}
We extract direct statistics from the numbers themselves, including consistency, repetition, monotonicity, sortedness, and decimal-precision behavior. These cues complement the observer-based mismatch by measuring whether the numeric channel behaves like genuine experimental reporting rather than a smooth synthetic pattern.

We denote the corresponding table-level feature blocks by $\phi_{\mathrm{struct}}(T)$ and $\phi_{\mathrm{num}}(T)$, and define the fused representation as
\begin{equation}
\phi_{\mathrm{FusedAll}}(T)=
\left[
\phi_{\mathrm{PPL4}}(T)\,;\,
\phi_{\mathrm{struct}}(T)\,;\,
\phi_{\mathrm{num}}(T)
\right].
\label{eq:fusedall}
\end{equation}
This design reflects the paper's main claim: the strongest evidence comes from combining a provenance-sensitive likelihood contrast with complementary structure and numeric regularities.

\subsection{From tables to a paper-level audit score}

As a paper may contain multiple tables with different sizes and roles, \textsc{TAB-AUDIT} aggregates table evidence at the paper level. For each feature dimension $k$, we use a robust median aggregation:
\begin{equation}
\Phi_p[k] = \mathrm{median}_{T \in \mathcal{T}_p}\, \phi_T[k],
\label{eq:paper_agg}
\end{equation}
where $\phi_T$ is either $\phi_{\mathrm{PPL4}}(T)$ or $\phi_{\mathrm{FusedAll}}(T)$ depending on the feature set under study.


The final output is a paper-level audit score
\[
s_p = h(\Phi_p),
\]
where $h$ is a standard detector backend. In the main signal-focused analysis, we use simple supervised backends to test whether the extracted features carry useful provenance information. Logistic regression serves as a transparent baseline because it makes the contribution of each signal easy to inspect. In a separate classifier-family comparison, we also evaluate MLP and Random Forest backends on the same paper-level features. Among the tested operational layers, Random Forest on \textsc{FusedAll} gives the strongest completed holdout result; however, backend choice is secondary to the main contribution, which is the paper-level signal representation itself.



\section{Experimental Setup}
\label{sec:experimental_setup}

We evaluate whether the forensic signals identified in Section~\ref{sec:case_study} translate into reliable \emph{paper-level} auditing. The main question is not only whether human and fabricated papers are separable in-domain, but also whether the same signal remains useful under a strict false-positive budget and under \emph{generator shift}. All experiments are conducted at the paper level, and final reporting separates in-domain evaluation from external holdout evaluation.

As a controlled realism diagnostic, we construct a \emph{prompt-only} comparison set in which a model is prompted directly with a topic and asked to generate a plausible experimental results table without first generating a full literature-grounded paper. Each generated table is written into the same extraction schema, with one generated table treated as one pseudo-paper. Because this diagnostic contains only 20 fabricated pseudo-papers, we use it as an auxiliary comparison and not as a replacement benchmark. The result can be find in Table.~\ref{tab:baselines_indomain}

\paragraph{Observer models and predictions}

Unless otherwise noted, the paper-level representations are the features defined in Section~\ref{sec:method}. Likelihood-based signals are computed by a frozen \emph{observer} language model that scores serialized table views but does not generate the papers being audited. We evaluate three observer families: GPT-2 \citep{radford2019gpt2}, Qwen3-8B \citep{yang2025qwen3}, and Llama3-8B \citep{grattafiori2024llama3}.


To keep the signal analysis primary, we use logistic regression as the default operational backend when comparing observer families. This gives a transparent and comparatively low-capacity detector, making it easier to interpret whether the features themselves are informative. We then compare logistic regression, Random Forest, and MLP only in a dedicated classifier-family comparison on the same paper-level features. In this framing, the classifier is a secondary operational layer, while the main object of study is the forensic signal.




\paragraph{Evaluation metrics}

False accusations against human-authored papers are costly, so we evaluate the detector as a \emph{screening tool} rather than as a generic balanced classifier. We focus on the 5\% operating point and therefore report \textbf{TPR@5\%} together with the corresponding \textbf{realized FPR@5\%}. These quantities directly answer the deployment-relevant question: under a small false-positive budget on human papers, how many fabricated papers are recovered, and does the calibrated threshold actually generalize to unseen data? We also report \textbf{AUROC} and \textbf{AUPRC} as threshold-free summaries of ranking quality. Under our framing, AUROC and AUPRC are supporting metrics, while the low-FPR results are the primary metrics.


\paragraph{Baselines and auxiliary comparisons}

We compare \textsc{TAB-AUDIT} against two external baselines adapted to serialized tables: \emph{Binoculars}, a zero-shot likelihood-ratio detector \citep{hans2024binoculars}, and a \emph{DetectGPT-style} perturbation baseline \citep{mitchell2023detectgpt}. Both are applied to table serializations and aggregated to the paper level.




\section{Results}
\label{sec::result}

\begin{table*}[t]

\centering
\small
\setlength{\tabcolsep}{4pt}
\resizebox{\textwidth}{!}{%
\begin{tabular}{llcccccccc}

\toprule
& & \multicolumn{4}{c}{GPT-4o In-Domain Test} & \multicolumn{4}{c}{GPT-5.2 External Holdout} \\
\cmidrule(lr){3-6} \cmidrule(lr){7-10}
Detector & Observer & AUROC & AUPRC & TPR@5\% & Real FPR@5\% & AUROC & AUPRC & TPR@5\% & Real FPR@5\% \\
\midrule
\multicolumn{10}{l}{\textbf{External Baselines}} \\
Binoculars (no training) & shared & 0.683 & 0.701 & 0.385 & 0.108 & -- & -- & -- & -- \\
DetectGPT-style (no training) & GPT-2 & 0.397 & 0.430 & 0.015 & 0.020 & -- & -- & -- & -- \\
\midrule
\multicolumn{10}{l}{\textbf{Observer Comparison: TAB-AUDIT (PPL4, LogReg)}} \\
TAB-AUDIT (PPL4, LogReg) & GPT-2 & 0.884 & 0.804 & 0.374 & 0.074 & 0.855 & 0.756 & 0.173 & 0.065 \\
TAB-AUDIT (PPL4, LogReg) & Qwen & 0.902 & 0.855 & 0.354 & 0.054 & 0.808 & 0.689 & 0.030 & 0.055 \\
TAB-AUDIT (PPL4, LogReg) & Llama & 0.866 & 0.820 & 0.246 & 0.039 & 0.881 & 0.793 & 0.096 & 0.040 \\
\midrule
\multicolumn{10}{l}{\textbf{Classifier Comparison on the Best-Transfer Observer Family}} \\
FusedAll (LogReg) & GPT-2 & 0.986 & 0.971 & 0.856 & 0.030 & 0.825 & 0.738 & 0.061 & 0.039 \\
FusedAll (RandomForest) & GPT-2 & 0.987 & 0.985 & 0.933 & 0.054 & 0.883 & 0.808 & 0.218 & 0.054 \\
FusedAll (MLP) & GPT-2 & 0.986 & 0.975 & 0.862 & 0.030 & 0.805 & 0.723 & 0.051 & 0.030 \\
\bottomrule
\end{tabular}%
}
\caption{main-results table for fabricated-table detection. The left block reports the fixed in-domain test split on human arXiv papers versus GPT-4o-fabricated papers. The right block reports transfer to the GPT-5.2 external holdout. This adjusted version removes the stricter 1\% operating-point columns and keeps the 5\% operating point together with threshold-free ranking metrics. Thresholds are calibrated on validation humans.}
\label{tab::main_short}
\end{table*}



Table~\ref{tab::main_short} should be read primarily as a test of whether the proposed forensic signals support reliable \emph{paper-level screening}, rather than as a generic classifier comparison.
\paragraph{The core mismatch signal is real.}
Even the compact PPL4 version of \textsc{TAB-AUDIT} already separates human and fabricated papers substantially better than external zero-shot baselines adapted from prose detection. With GPT-2, Qwen, and Llama observers, \textsc{TAB-AUDIT} (PPL4, LogReg) reaches AUROC/AUPRC of 0.884/0.804, 0.902/0.855, and 0.866/0.820, respectively, whereas Binoculars reaches 0.683/0.701 and the DetectGPT-style baseline drops to 0.397/0.430. This gap supports the paper's main claim that provenance cues for scientific tables are not well captured by off-the-shelf detectors designed for free-form text. The useful evidence here is not generic fluency, but the mismatch between a table's scientific scaffold and its numeric behavior.

\paragraph{Observer choice matters, but in-domain ranking and transfer are not aligned.}
Among the PPL4 observers, Qwen gives the strongest in-domain ranking metrics (AUROC 0.902, AUPRC 0.855), but it is the weakest under generator shift at the deployment-relevant operating point, reaching only TPR@5\%=0.030 on the GPT-5.2 holdout. By contrast, Llama yields the strongest holdout ranking metrics (AUROC 0.881, AUPRC 0.793), while GPT-2 gives the strongest holdout low-FPR recall (TPR@5\%=0.173). This split is important for interpretation: a detector can rank papers well overall without being the best choice for strict screening. In other words, threshold-free separability and low-FPR recovery should be treated as related but distinct properties.


\paragraph{\textsc{FusedAll} with Random Forest is the strongest operational backend.}
Once the fused signal representation is fixed, the classifier-family comparison shows that Random Forest is the strongest completed backend among the tested operational layers. In-domain, \textsc{FusedAll} + Random Forest reaches AUROC 0.987, AUPRC 0.985, and TPR@5\%=0.933. On the GPT-5.2 holdout, it remains the best completed system with AUROC 0.883, AUPRC 0.808, and TPR@5\%=0.218, outperforming the logistic and MLP variants at the same feature level. We therefore treat Random Forest as the strongest \emph{deployment-oriented} backend, while keeping the paper's main scientific contribution on the signal representation itself: fabricated papers can imitate scientific skeletons more easily than they can reproduce the distributional texture of genuinely measured results.

\begin{figure}[t]
  \centering
  \includegraphics[width=\columnwidth]{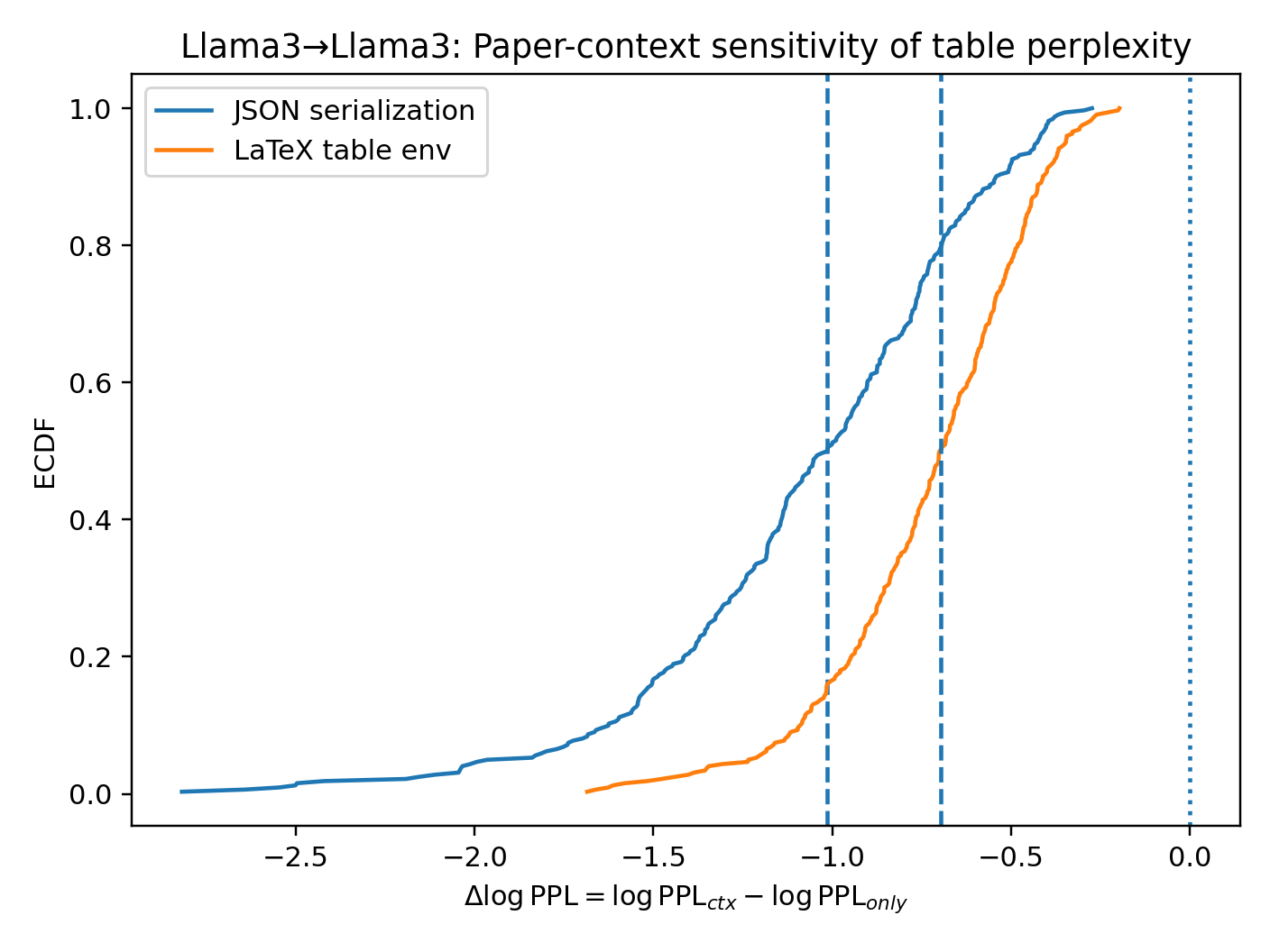}
  \caption{
    \textbf{Paper-context sensitivity of table perplexity.}
    Empirical CDF of $\Delta \log \mathrm{PPL} = \log \mathrm{PPL}_{ctx}-\log \mathrm{PPL}_{only}$,
    where $\mathrm{PPL}_{ctx}$ scores table tokens conditioned on the preceding paper content
    (prefix masked from loss), and $\mathrm{PPL}_{only}$ scores the same table tokens in isolation.
    Negative values indicate that paper context makes the table more predictable under the observer.
  }
  \label{fig:ctx-ablation-ecdf}
\end{figure}

\section{Related Work}

\paragraph{Scientific-integrity forensics for reported results.}
Before LLM-generated papers became a practical concern, fabricated scientific evidence was often approached through forensic checks on arithmetic constraints, digit regularities, and internal consistency of reported statistics. Examples include the GRIM family of granularity checks for discrete measurements \citep{brown2017grim,anaya2016grimmer} and automated verification of reported test statistics and $p$-values via \texttt{statcheck} \citep{nuijten2016statcheck}. These tools are powerful when their assumptions hold, but they are not designed for a regime in which an entire paper---including its tables---can be synthesized by an LLM.

\paragraph{Fabricated papers and machine-generated text.}
The broader integrity literature has documented systematic manipulation at scale, including duplicated images, templated text, and paper-mill behavior \citep{bik2016duplication,byrne2024papermills}. In parallel, post-hoc LLM detection has largely focused on free-form text: DetectGPT uses probability curvature under perturbations \citep{mitchell2023detectgpt}, Fast-DetectGPT reduces cost through conditional sampling \citep{bao2024fastdetectgpt}, and Binoculars contrasts two related LLMs for zero-shot detection \citep{hans2024binoculars}. A recurring theme is that performance can degrade under domain shift and human editing, which is precisely why our evaluation emphasizes paper-level aggregation, holdout-generator transfer, and low-FPR calibration.


\section{Conclusion}
We introduce the dataset \ourdata and the \textit{first} systematic study on detecting AI-generated fabricated tables in scientific manuscripts and demonstrate that table-centric signals provide a reliable basis for identifying fabricated research. Our framework, \ourmethod, leverages the mismatch between textual structure and numerical content, enabling effective paper-level auditing without access to underlying experimental artifacts. Empirical results show that \textsc{TAB-AUDIT} achieves high identification rates with low false positive rates on both in-domain and out-of-domain settings on \ourdata. These findings support table-centric, paper-level auditing as a practical and scalable tool for surfacing high-risk papers for manual review.

\section*{Limitations}

TAB-AUDIT depends on both the observer language model and the table extraction/serialization pipeline. As our experiments suggest, highly aligned instruction-tuned observers can smooth over inconsistencies and become weaker auditors than older base models. Performance may also degrade on papers with complex PDF layouts, merged cells, figure-like tables, or extraction errors that distort the serialized views. The serialization also discards some visual and formatting cues that may be informative to human reviewers. Because the method operates only on extracted table artifacts, it cannot use evidence from code, logs, raw outputs, or richer paper context.
Although the method transfers to held-out generators, the gap between in-domain and external-holdout results shows that robustness to generator shift remains incomplete. Our paper-level median aggregation is robust to noisy tables, but it can dilute evidence when only one table in an otherwise normal paper is suspicious. We therefore view TAB-AUDIT as an artifact-only, low-FPR screening tool for prioritizing manual review, not as standalone evidence of fabrication or misconduct. Screening outputs should not be interpreted as prevalence estimates or allegations about specific papers.


\bibliography{custom}

\newpage
\appendix
\section*{Appendix}
\section{Likelihood Mismatch as Evidence of Process Shift}
\label{sec:theory}

This appendix gives a compact formal interpretation of why multi-view likelihood mismatch can separate human experimental tables from LLM-fabricated tables.

\subsection{Two-channel view of tables}
We idealize a table $T$ as consisting of two coupled channels:
a \emph{skeleton} $A$ (titles, headers, row labels, layout tokens) and
a \emph{numeric evidence channel} $B$ (numbers, represented in our implementation through digit-spaced serialization).
Human papers induce a distribution $P_H(A,B)$, while AI-fabricated papers induce $P_A(A,B)$.

A key empirical hypothesis motivating our method is that $P_A(A)$ can be closer to $P_H(A)$ than $P_A(B \mid A)$ is to $P_H(B \mid A)$. Intuitively, modern LLMs are strong at producing conventional table skeletons, but the numbers must still be invented in a way that imitates an experimental process.

\subsection{Within-table likelihood mismatch}
\label{sec:theory_mismatch}
Recall the observer LM $q$ and the sequence score $\mathrm{NLL}_q(\cdot)$ (equivalently $\log\mathrm{PPL}_q(\cdot)$) defined in Section~\ref{sec:method} (Eq.~\ref{nll}). Given a table $T$, we form two views:
(i) a \emph{skeleton} string $s^{\text{text}}(T)$ (title, headers, notes, and layout with numbers masked),
and (ii) a \emph{digit-spaced numeric} string $s^{\text{dig}}(T)$.
We define the mismatch score as
\begin{equation}
\label{eq:mismatch}
m_q(T) \,=\, \mathrm{NLL}_q\!\left(s^{\text{dig}}(T)\right) \, - \, \mathrm{NLL}_q\!\left(s^{\text{text}}(T)\right).
\end{equation}

\paragraph{Key hypothesis.}
Human experimental tables are produced by measurement and reporting conventions, whereas fabricated tables are produced by conditional text sampling. This process difference induces a systematic shift in the \emph{relative} surprisal of numbers versus skeleton:
\begin{equation}
\label{eq:mismatch_gap}
\mathbb{E}\!\left[m_q(T)\mid \textsc{AI}\right] \,>
\mathbb{E}\!\left[m_q(T)\mid \textsc{Human}\right].
\end{equation}
Eq.~\ref{eq:mismatch_gap} motivates thresholding or learning on $m_q(T)$ and its paper-level aggregates.

\paragraph{Paper-level score.}
A paper $p$ contains multiple tables $\{T_j\}_{j=1}^{K_p}$; we aggregate table evidence robustly via
\begin{equation}
\label{eq:paper_mismatch}
M_q(p) \,=\, \mathrm{median}_{j\in[K_p]} \, m_q(T_j).
\end{equation}
Because $m_q(T)$ compares two views of the \emph{same} table, it cancels many shared confounds such as topic, domain vocabulary, and broad stylistic packaging. This helps explain why the signal transfers more gracefully than detectors built only on surface form.

\section{Full main table result}

\begin{table*}[t]
\centering
\scriptsize
\setlength{\tabcolsep}{4pt}
\caption{Full in-domain combination matrix from \texttt{exp105\_comprehensive\_main\_table}. All rows are evaluated on the same fixed paper-level test split of human arXiv papers versus GPT-4o-fabricated papers. Thresholds are calibrated on validation humans. This adjusted version reports the 5\% operating point rather than the stricter 1\% operating point.}
\label{tab:appendix-full-in-domain-combinations}
\begin{tabular}{llccccc}
\toprule
Observer & Baseline & AUROC & AUPRC & TPR@5\% & Real FPR@5\% & F1 \\
\midrule
gpt2 & DigitLaw+Precision(LogReg) & 0.950 & 0.926 & 0.703 & 0.054 & 0.888 \\
gpt2 & DigitsOnlyPPL(LogReg) & 0.890 & 0.818 & 0.256 & 0.039 & 0.838 \\
gpt2 & FusedAll(PPL4+Struct+DigitLaw, LogReg) & 0.986 & 0.971 & 0.856 & 0.030 & 0.955 \\
gpt2 & MismatchOnly(LogReg) & 0.637 & 0.638 & 0.144 & 0.044 & 0.638 \\
gpt2 & Structure+NumCellDensity(LogReg) & 0.969 & 0.939 & 0.882 & 0.044 & 0.944 \\
gpt2 & StructureOnly(LogReg) & 0.963 & 0.915 & 0.851 & 0.049 & 0.937 \\
gpt2 & TAB-AUDIT(PPL4, LogReg) & 0.884 & 0.804 & 0.374 & 0.074 & 0.842 \\
gpt2 & TextOnlyPPL(LogReg) & 0.850 & 0.751 & 0.185 & 0.064 & 0.830 \\
gpt2 & DetectGPT-style (no training) & 0.397 & 0.430 & 0.015 & 0.020 & 0.637 \\
gpt2 & FusedAll(PPL4+Struct+DigitLaw, MLP) & 0.986 & 0.975 & 0.862 & 0.030 & 0.967 \\
gpt2 & FusedAll(PPL4+Struct+DigitLaw, RandomForest) & 0.987 & 0.985 & 0.933 & 0.054 & 0.948 \\
gpt2 & TAB-AUDIT(PPL4, MLP) & 0.923 & 0.886 & 0.369 & 0.025 & 0.851 \\
gpt2 & TAB-AUDIT(PPL4, RandomForest) & 0.923 & 0.900 & 0.267 & 0.020 & 0.867 \\
llama & DigitLaw+Precision(LogReg) & 0.950 & 0.926 & 0.703 & 0.054 & 0.888 \\
llama & DigitsOnlyPPL(LogReg) & 0.863 & 0.769 & 0.185 & 0.054 & 0.815 \\
llama & FusedAll(PPL4+Struct+DigitLaw, LogReg) & 0.983 & 0.969 & 0.867 & 0.039 & 0.958 \\
llama & MismatchOnly(LogReg) & 0.756 & 0.737 & 0.159 & 0.039 & 0.699 \\
llama & Structure+NumCellDensity(LogReg) & 0.969 & 0.939 & 0.882 & 0.044 & 0.944 \\
llama & StructureOnly(LogReg) & 0.963 & 0.915 & 0.851 & 0.049 & 0.937 \\
llama & TAB-AUDIT(PPL4, LogReg) & 0.866 & 0.820 & 0.246 & 0.039 & 0.807 \\
llama & TextOnlyPPL(LogReg) & 0.838 & 0.762 & 0.169 & 0.064 & 0.767 \\
llama & FusedAll(PPL4+Struct+DigitLaw, MLP) & 0.986 & 0.972 & 0.851 & 0.025 & 0.959 \\
llama & FusedAll(PPL4+Struct+DigitLaw, RandomForest) & 0.988 & 0.987 & 0.949 & 0.054 & 0.949 \\
llama & TAB-AUDIT(PPL4, MLP) & 0.909 & 0.880 & 0.195 & 0.025 & 0.858 \\
llama & TAB-AUDIT(PPL4, RandomForest) & 0.911 & 0.886 & 0.256 & 0.025 & 0.844 \\
qwen & DigitLaw+Precision(LogReg) & 0.950 & 0.926 & 0.703 & 0.054 & 0.888 \\
qwen & DigitsOnlyPPL(LogReg) & 0.869 & 0.800 & 0.236 & 0.044 & 0.821 \\
qwen & FusedAll(PPL4+Struct+DigitLaw, LogReg) & 0.984 & 0.969 & 0.882 & 0.030 & 0.946 \\
qwen & MismatchOnly(LogReg) & 0.794 & 0.772 & 0.210 & 0.054 & 0.734 \\
qwen & Structure+NumCellDensity(LogReg) & 0.969 & 0.939 & 0.882 & 0.044 & 0.944 \\
qwen & StructureOnly(LogReg) & 0.963 & 0.915 & 0.851 & 0.049 & 0.937 \\
qwen & TAB-AUDIT(PPL4, LogReg) & 0.902 & 0.855 & 0.354 & 0.054 & 0.855 \\
qwen & TextOnlyPPL(LogReg) & 0.853 & 0.782 & 0.164 & 0.044 & 0.815 \\
qwen & FusedAll(PPL4+Struct+DigitLaw, MLP) & 0.977 & 0.962 & 0.764 & 0.025 & 0.934 \\
qwen & FusedAll(PPL4+Struct+DigitLaw, RandomForest) & 0.987 & 0.984 & 0.938 & 0.049 & 0.957 \\
qwen & TAB-AUDIT(PPL4, MLP) & 0.913 & 0.869 & 0.328 & 0.044 & 0.848 \\
qwen & TAB-AUDIT(PPL4, RandomForest) & 0.912 & 0.877 & 0.308 & 0.025 & 0.859 \\
shared & Binoculars (no training) & 0.683 & 0.701 & 0.385 & 0.108 & 0.656 \\
\bottomrule
\end{tabular}
\end{table*}

\begin{table*}[t]
\centering
\small
\setlength{\tabcolsep}{4.5pt}
\begin{tabular}{llccccc}
\toprule
Detector & Observer & AUROC & AUPRC & TPR@5\% & Real FPR@5\% & F1 \\
\midrule
\multicolumn{7}{l}{\textbf{GPT-5.2 External Holdout}} \\
TAB-AUDIT(PPL4, LogReg) & GPT-2 & 0.855 & 0.756 & 0.173 & 0.065 & -- \\
TAB-AUDIT(PPL4, LogReg) & Qwen & 0.808 & 0.689 & 0.030 & 0.055 & -- \\
TAB-AUDIT(PPL4, LogReg) & Llama & 0.881 & 0.793 & 0.096 & 0.040 & -- \\
FusedAll(LogReg) & GPT-2 & 0.825 & 0.738 & 0.061 & 0.039 & 0.401 \\
FusedAll(RandomForest) & GPT-2 & 0.883 & 0.808 & 0.218 & 0.054 & 0.641 \\
FusedAll(MLP) & GPT-2 & 0.805 & 0.723 & 0.051 & 0.030 & 0.216 \\
\bottomrule
\end{tabular}
\caption{Supplementary external-holdout transfer results. Holdout metrics are currently available for the canonical TAB-AUDIT observer comparison and for the GPT-2-based fused classifier-family comparison. This adjusted version reports the 5\% operating point rather than the stricter 1\% operating point. Dashes indicate that a directly comparable holdout run was not produced in the current experiment set.}
\label{tab:appendix-holdout-combinations}
\end{table*}

We present full results of all combination of detectors methods here as Table .~\ref{tab:appendix-full-in-domain-combinations}.

\section{Dataset Details and Statistics}
\label{app:dataset_stats}

Table~\ref{tab:dataset_full} summarizes the benchmark counts used in the current paper. We report only the counts that are directly supported by the current experiment artifacts: paper totals for each split, total table counts where available, and the auxiliary evaluation sets used in the completed extension experiments. The broader server-side corpus inventory is larger, but the paper's benchmark claims rely only on the filtered human corpus, the filtered GPT-4o fabricated-paper corpus, the GPT-5.2 holdout corpus, and the prompt-only comparison set.

\begin{table*}[t]
\centering
\small
\caption{Current benchmark and evaluation-set counts used in the paper. Split-level table counts are omitted where they are not finalized in the saved artifacts.}
\label{tab:dataset_full}
\begin{tabular}{lcccc}
\toprule
\textbf{Subset} & \textbf{Human} & \textbf{AI} & \textbf{Total Papers} & \textbf{Total Tables} \\
\midrule
Clean benchmark & 1012 & 976 & 1988 & 5284 \\
Train split & 758 & 732 & 1490 & -- \\
Validation split & 51 & 49 & 100 & -- \\
Test split & 203 & 195 & 398 & -- \\
GPT-5.2 holdout & 203 & 197 & 400 & 1324 \\
Prompt-only baseline & 1012 & 20 & 1032 & 3274 \\
\bottomrule
\end{tabular}
\end{table*}

Table~\ref{tab:benchmark_filtering} makes the benchmark filtering step explicit. The saved clean-benchmark report starts from 2111 indexed papers and removes 123 cases during validity filtering before the final train/validation/test split is formed.

\begin{table}[t]
\centering
\small
\caption{Filtering from the indexed source pool to the final clean benchmark.}
\label{tab:benchmark_filtering}
\begin{tabular}{lrrr}
\toprule
\textbf{Stage} & \textbf{Human} & \textbf{AI} & \textbf{Total} \\
\midrule
Indexed source pool & 1118 & 993 & 2111 \\
Dropped during filtering & 106 & 17 & 123 \\
Final usable benchmark & 1012 & 976 & 1988 \\
\bottomrule
\end{tabular}
\end{table}

The saved GPT-5.2 holdout report contains 203 human papers and 197 AI papers, corresponding to 641 human tables and 683 GPT-5.2 tables. The saved prompt-only comparison artifacts contain 1032 valid papers and 3274 extracted tables in total, split into 824 training papers, 104 validation papers, and 104 test papers. These appendix counts are included to document corpus composition, not to redefine the main benchmark.
\begin{table*}[t]
\centering
\small
\begin{tabular}{lcccccc}
\toprule
Metric & Human & GPT4o & Qwen3-8B & Llama3.1-8B & min$|d|$ & mean$|d|$ \\
\midrule
Row $\Delta$-consistency  & \textbf{0.497$\pm$0.433} & 0.235$\pm$0.390 & 0.187$\pm$0.359 & 0.206$\pm$0.367 & 0.64 & 0.71 \\
Most-common value rate &  \textbf{0.113$\pm$0.138} & 0.199$\pm$0.136 & 0.229$\pm$0.153 & 0.220$\pm$0.153 & 0.63 & 0.72 \\
Strict monotone-column rate & 0.190$\pm$0.324 & 0.418$\pm$0.438 & 0.445$\pm$0.461 & 0.560$\pm$0.474 & 0.59 & 0.71 \\
Mean Spearman trend strength & 0.558$\pm$0.262 & 0.704$\pm$0.244 & 0.733$\pm$0.256 & 0.770$\pm$0.267 & 0.58 & 0.68 \\
Row \%$\Delta$-consistency & 0.357$\pm$0.374 & 0.169$\pm$0.339 & 0.153$\pm$0.296 & 0.150$\pm$0.315 & 0.53 & 0.58 \\
Mean column sortedness & 0.754$\pm$0.133 & 0.811$\pm$0.150 & 0.831$\pm$0.154 & 0.864$\pm$0.154 & 0.40 & 0.57 \\
Dec\_places\_mean & \textbf{1.589$\pm$1.045} & 0.987$\pm$0.572 & 0.780$\pm$0.692 & 1.247$\pm$0.642 & 0.39 & 0.67 \\
Adjacent direction consistency & 0.711$\pm$0.136 & 0.740$\pm$0.202 & 0.770$\pm$0.202 & 0.808$\pm$0.214 & 0.17 & 0.35 \\
\bottomrule
\end{tabular}%

\caption{Top LM-free statistical signals (mean$\pm$std over tables).
Row $\Delta$-consistency checks whether any triple in a row satisfies $c\approx b-a$; Row \%$\Delta$-consistency checks $c\approx (b-a)/|a|\times100$.
Most-common value rate measures numeric repetition. Monotonicity/sortedness quantify over-regular trends.
min$|d|$/mean$|d|$ report absolute Cohen's $d$ vs.\ human, aggregated across generators.}
\label{tab:top_stats:wide}
\end{table*}

\section{LM-Free Numeric Table Statistics}
\label{app:lmfree-stats}
The main paper now reports the LM-free numeric statistics in Section~\ref{sec:case_study} (Table~\ref{tab:top_stats:wide}) because they directly motivate the paper's signal analysis.
From the Table.~\ref{tab:top_stats:wide}, fabricated tables are markedly more \emph{over-regular}. 
They have higher most-common-value rates, stronger monotone-column behavior, stronger trend strength, and higher sortedness. 
This does not mean that fabricated tables are obviously fake; on the contrary, they often look neat and locally plausible. 
But they are too neat in a specific way: generation tends to drift toward repeated values, tidy progressions, and smoother ordinal structure than is typical in real experimental reporting.

Third, human tables exhibit richer \emph{precision behavior}. 
The mean number of decimal places is higher in human tables, whereas fabricated tables more often collapse toward a narrower and more templated precision profile. 
This again matches the broader hypothesis: genuine experiments produce irregular numeric traces, while synthesized tables often converge to a cleaner and more homogeneous style of number writing.

An important detail is that Table~\ref{tab:top_stats:wide} reports both the minimum and mean absolute Cohen's $d$ against the human distribution across multiple generator families. 
Many of the strongest LM-free signals remain sizeable even under the most conservative cross-generator comparison, which suggests that these effects are not just quirks of a single model. 
They instead point to a broader process-level difference between experiment-grounded and fabricated numeric reporting.





\section{Full generation pipeline}
\begin{figure*}[t]
    \centering
    \includegraphics[width=0.8\linewidth]{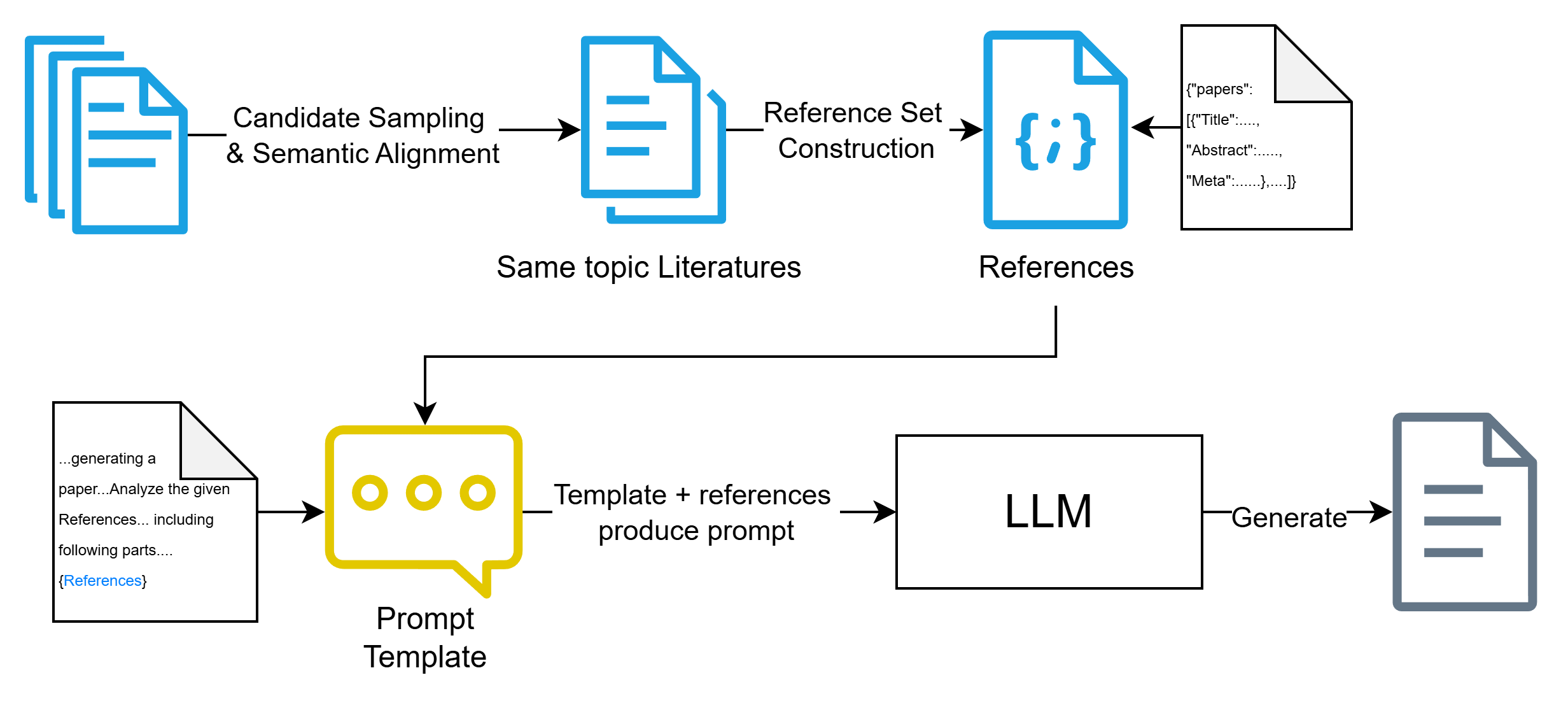}
    \caption{Literature-grounded AI paper generation pipeline used to construct the fabricated-paper benchmark.}
    \label{fig:generation_pipeline}
\end{figure*}
Table~\ref{fig:generation_pipeline} shows the full generation pipeline we described.
\subsection{Prompt Template for Paper Generation}
\label{sec:prompt}

We use a structured prompt to generate fully formed scientific papers from a language model given a set of reference abstracts. This prompt serves as a controlled mechanism for producing fabricated scientific content with realistic academic structure, enabling systematic analysis and evaluation.

\begin{quote}
\small
\textbf{Paper Generation Prompt.}

You are a research assistant AI tasked with generating a complete scientific paper based on provided literature. You should:

\begin{enumerate}
    \item Analyze the given references.
    \item Identify gaps in existing research to motivate a new study.
    \item Propose a coherent main research idea.
    \item Generate the full paper in \LaTeX{} format, including the following sections:
    Title, Abstract, Introduction, Related Work, Methods/Experiment, Results (with tables), Discussion, Conclusion, and Contributions.
\end{enumerate}

All generated content must be original, academically rigorous, and follow standard scientific writing conventions. The model is instructed to ground the paper in the provided references and integrate them naturally.

\medskip
\textbf{Input:} References with abstracts.
\end{quote}

\begin{figure*}[t]
    \centering
    \includegraphics[width=0.7\linewidth]{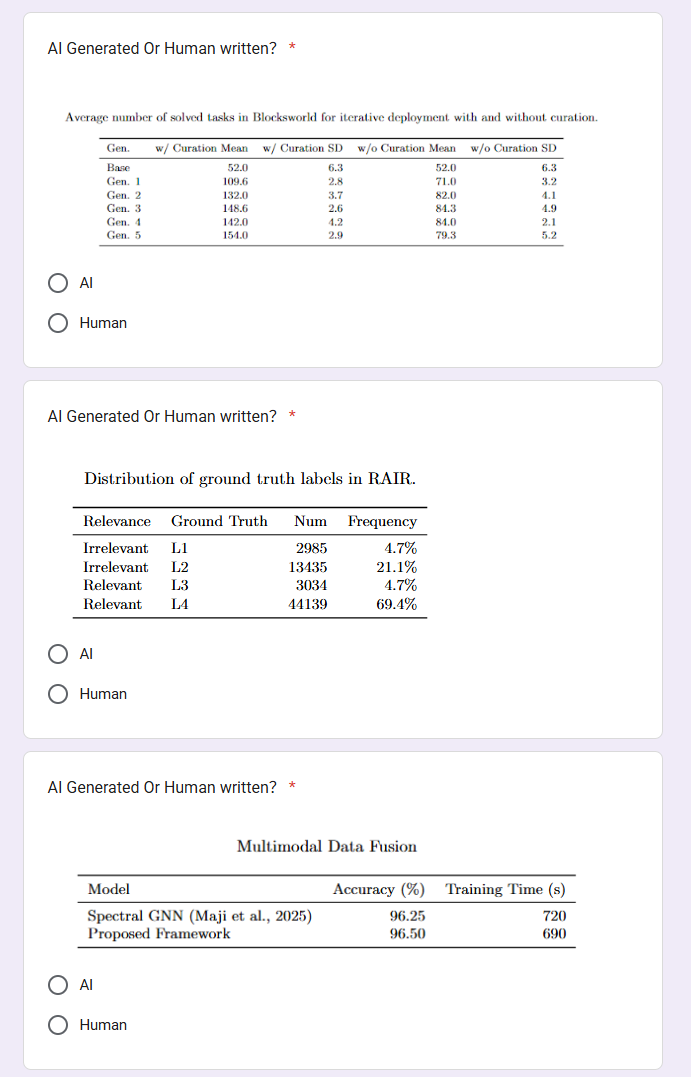}
    \caption{Table Authenticity Judgment Form.}
    \label{fig:from_screenshot}
\end{figure*}
\section{Train-Source and Observer Robustness}
\label{app:train_source_robustness}

This supplementary experiment isolates the effect of the AI training source from the effect of the observer model while keeping the detector family fixed to the canonical \textsc{TAB-AUDIT}(PPL4, LogReg) configuration. We train on size-matched fabricated-paper sets from GPT-4o, Llama3.1, and Qwen2.5, and score the resulting table views with three observers: GPT-2, Qwen, and Llama. Table~\ref{tab:train_source_robustness_full} reports the full cross-product over train source, observer, and test generator.

The matrix shows that robustness is not explained by a single matched-family rule. Qwen2.5 training is the strongest source on three of the four evaluation sets by AUROC, but the GPT-5.2 holdout instead prefers GPT-4o training with a Llama observer. This pattern supports the paper's broader interpretation: the auditing signal is transferable, but its strongest operational realization depends on the particular generator shift being faced.

\begin{table*}[t]
\centering
\small
\resizebox{\textwidth}{!}{%
\begin{tabular}{lllrrrr}
\toprule
Test Set & Train Source & Observer & AUROC & AUPRC & TPR@5\% & Real FPR@5\% \\
\midrule

\multicolumn{7}{l}{\textbf{In-domain (GPT-4o test)}} \\
in-domain & GPT-4o & GPT-2 & 0.938 & 0.936 & 0.723 & 0.064 \\
in-domain & Llama3.1 & GPT-2 & 0.902 & 0.911 & 0.596 & 0.043 \\
in-domain & Qwen2.5 & GPT-2 & 0.925 & 0.930 & 0.638 & 0.043 \\
in-domain & GPT-4o & Qwen & 0.933 & 0.932 & 0.894 & 0.170 \\
in-domain & Llama3.1 & Qwen & 0.925 & 0.918 & 0.830 & 0.170 \\
\textbf{in-domain} & \textbf{Qwen2.5} & \textbf{Qwen} & \textbf{0.981} & \textbf{0.975} & \textbf{1.000} & \textbf{0.106} \\
in-domain & GPT-4o & Llama & 0.893 & 0.871 & 0.809 & 0.149 \\
in-domain & Llama3.1 & Llama & 0.932 & 0.928 & 0.787 & 0.085 \\
in-domain & Qwen2.5 & Llama & 0.961 & 0.956 & 0.872 & 0.085 \\

\midrule
\multicolumn{7}{l}{\textbf{GPT-5.2 holdout}} \\
GPT-5.2 & GPT-4o & GPT-2 & 0.901 & 0.966 & 0.452 & 0.064 \\
GPT-5.2 & Llama3.1 & GPT-2 & 0.931 & 0.980 & 0.635 & 0.043 \\
GPT-5.2 & Qwen2.5 & GPT-2 & 0.886 & 0.961 & 0.386 & 0.043 \\
GPT-5.2 & GPT-4o & Qwen & 0.883 & 0.951 & 0.660 & 0.170 \\
GPT-5.2 & Llama3.1 & Qwen & 0.764 & 0.901 & 0.447 & 0.170 \\
GPT-5.2 & Qwen2.5 & Qwen & 0.638 & 0.854 & 0.147 & 0.106 \\
\textbf{GPT-5.2} & \textbf{GPT-4o} & \textbf{Llama} & \textbf{0.937} & \textbf{0.977} & \textbf{0.898} & \textbf{0.149} \\
GPT-5.2 & Llama3.1 & Llama & 0.861 & 0.934 & 0.416 & 0.085 \\
GPT-5.2 & Qwen2.5 & Llama & 0.825 & 0.921 & 0.234 & 0.085 \\

\midrule
\multicolumn{7}{l}{\textbf{Llama3.1 holdout}} \\
Llama3.1 & GPT-4o & GPT-2 & 0.785 & 0.947 & 0.380 & 0.064 \\
Llama3.1 & Llama3.1 & GPT-2 & 0.913 & 0.980 & 0.650 & 0.043 \\
Llama3.1 & Qwen2.5 & GPT-2 & 0.886 & 0.974 & 0.538 & 0.043 \\
Llama3.1 & GPT-4o & Qwen & 0.846 & 0.963 & 0.620 & 0.170 \\
Llama3.1 & Llama3.1 & Qwen & 0.935 & 0.983 & 0.868 & 0.170 \\
Llama3.1 & Qwen2.5 & Qwen & 0.930 & 0.982 & 0.850 & 0.106 \\
Llama3.1 & GPT-4o & Llama & 0.772 & 0.935 & 0.440 & 0.149 \\
Llama3.1 & Llama3.1 & Llama & 0.933 & 0.983 & 0.774 & 0.085 \\
\textbf{Llama3.1} & \textbf{Qwen2.5} & \textbf{Llama} & \textbf{0.938} & \textbf{0.985} & \textbf{0.778} & \textbf{0.085} \\

\midrule
\multicolumn{7}{l}{\textbf{Qwen2.5 holdout}} \\
Qwen2.5 & GPT-4o & GPT-2 & 0.847 & 0.975 & 0.489 & 0.064 \\
Qwen2.5 & Llama3.1 & GPT-2 & 0.888 & 0.983 & 0.603 & 0.043 \\
Qwen2.5 & Qwen2.5 & GPT-2 & 0.923 & 0.988 & 0.680 & 0.043 \\
Qwen2.5 & GPT-4o & Qwen & 0.911 & 0.986 & 0.786 & 0.170 \\
Qwen2.5 & Llama3.1 & Qwen & 0.964 & 0.994 & 0.949 & 0.170 \\
\textbf{Qwen2.5} & \textbf{Qwen2.5} & \textbf{Qwen} & \textbf{0.973} & \textbf{0.996} & \textbf{0.954} & \textbf{0.106} \\
Qwen2.5 & GPT-4o & Llama & 0.745 & 0.948 & 0.371 & 0.149 \\
Qwen2.5 & Llama3.1 & Llama & 0.934 & 0.989 & 0.774 & 0.085 \\
Qwen2.5 & Qwen2.5 & Llama & 0.960 & 0.994 & 0.857 & 0.085 \\

\bottomrule
\end{tabular}%
}
\caption{Full train-source/observer robustness matrix for the canonical \textsc{TAB-AUDIT}(PPL4, LogReg) detector. Rows vary the fabricated-paper source used for supervised training and the observer model used for likelihood scoring. Best AUROC row within each test set is boldfaced.}
\label{tab:train_source_robustness_full}
\end{table*}

\section{Inspecting ACL 2024-2025 Proceedings}

\label{sec:acl-screening}

We further apply the fabricated-table detector as a screening tool over recent ACL proceedings. This experiment is not treated as a labeled benchmark, since ground-truth fabrication labels are unavailable. Instead, the detector is used only to rank papers by risk and to identify a small set of high-scoring cases for manual review.

Using the extracted-table pipeline over the ACL 2024 and 2025 main proceedings, we obtained table extractions for 1,469 papers, comprising 4,473 extracted tables in total. We then re-scored this extracted corpus with the strongest completed detector from our classifier-family comparison, namely \texttt{FusedAll + RandomForest}. Under this detector, all 1,469 extracted papers received scores. The resulting score distribution has mean 0.115, standard deviation 0.123, median 0.069, and maximum 0.917, indicating that most papers receive relatively low detector scores while a small tail remains substantially higher risk.

Following the same validation-human calibration protocol used elsewhere in the paper, we applied the stored 5\% FPR operating threshold from the corresponding in-domain model, which yielded a screening threshold of 0.5964. Under this threshold, 17 ACL papers were flagged as high-risk, corresponding to 1.16\% of scored papers.

These outputs should be interpreted strictly as triage signals rather than evidence of misconduct. A high score may reflect unusual table structure, extraction noise, atypical numeric formatting, or detector mismatch under domain shift. Accordingly, the role of this experiment is to estimate the prevalence of detector-flagged papers and to prioritize a manageable subset of papers for manual review.

\section{Train-source and observer robustness.}
To disentangle robustness to the observer model from robustness to the AI training source, we keep the canonical \textsc{TAB-AUDIT}(PPL4, LogReg) detector fixed and vary two axes: the fabricated-paper source used for supervised training (GPT-4o, Qwen2.5, or Llama3.1) and the observer model used for likelihood scoring (GPT-2, Qwen, or Llama). From the Appendix~\ref{app:train_source_robustness}, two findings stand out. First, no single observer--training-source pair dominates all regimes: the in-domain benchmark and the Qwen2.5 holdout are best with Qwen2.5 training and a Qwen observer, whereas the Llama3.1 holdout is best with Qwen2.5 training and a Llama observer. Second, the GPT-5.2 holdout behaves differently: its best row is obtained by training on GPT-4o fabrications and scoring with Llama. This indicates that the provenance signal transfers across generator families, but the strongest operational pairing depends on the target shift rather than following a simple matched-family rule.

\begin{table*}[ht]
\centering
\small
\resizebox{\textwidth}{!}{%
\begin{tabular}{p{4.0cm}p{2.6cm}cccp{4.2cm}}
\toprule
\textbf{Experiment} & \textbf{Eval. set} & \textbf{Papers} & \textbf{AUROC} & \textbf{AUPRC} & \textbf{Comment} \\
\midrule
GPT5.2 Holdout set & Clean test split & 398 & 0.8857 & 0.8093 & 203 human / 195 AI papers \\
Prompt-only baseline & Prompt-only test split & 796 & 0.973 & 0.886 &  398 Human vs  398 prompt-only pseudo-papers \\
\bottomrule
\end{tabular}
}
\caption{Completed paper-level evaluations that do not involve generator-shift transfer.}
\label{tab:baselines_indomain}
\end{table*}


\section{Benchmark splits}
\label{sec:benchmark_splits}

All splits are formed at the \emph{paper} level. Each paper is represented by an extracted-table JSON file, and no paper contributes tables to more than one partition. This grouped protocol prevents leakage across tables from the same manuscript.

The primary labeled benchmark combines the filtered recent-human corpus with the filtered GPT-4o fabricated-paper corpus produced by the literature-grounded pipeline in Section~\ref{sec:generation} \citep{openai2024gpt4o}. The indexed source pool contains 2,111 candidate papers in total: 1,118 human papers and 993 fabricated papers. After removing failed or unusable extractions, the clean benchmark contains 1,012 human papers and 976 fabricated papers, for 1,988 papers and 5,284 extracted tables in total. We use a fixed split of 1,490 training papers (758 human, 732 fabricated), 100 validation papers (51 human, 49 fabricated), and 398 test papers (203 human, 195 fabricated).


\section{Running Examples Across Human and Synthetic Sources}
\label{sec:appendix-running-examples}

To make the benchmark more concrete, we include one running example from each of four sources: a real human paper, a GPT-4o fabricated paper from the main synthetic benchmark, a GPT-5.2 fabricated paper from the external holdout set, and a prompt-only generated table. These examples are chosen to illustrate the qualitative differences that motivate our experiments rather than to serve as exhaustive case studies. The selected paper-level detector scores are 0.0995 for the human paper, 0.9992 for the GPT-4o synthetic paper, 0.7504 for the GPT-5.2 holdout paper, and 0.5231 for the prompt-only sample.

The four examples differ in ways that align with the benchmark results. The human table in Table~\ref{tab:running-human} is embedded in a broader empirical paper and reflects a genuine evaluation structure with multiple systems, multiple languages, and non-uniform performance patterns. The GPT-4o example in Table~\ref{tab:running-gpt4o} is much more stylized: it presents a generic ``Proposed Framework'' that cleanly dominates a baseline in a compact comparison table. The GPT-5.2 example in Table~\ref{tab:running-gpt52} is more elaborate than the prompt-only case, with multiple controlled conditions, averages, and a polished explanatory note, making it a stronger holdout example than direct table prompting alone. Finally, the prompt-only example in Table~\ref{tab:running-prompt} is plausible but self-contained: it resembles a benchmark summary table without the broader paper-level context or multi-table consistency pattern found in the full synthetic-paper pipelines.

Taken together, these running examples help explain the main empirical findings of the paper. Prompt-only tables are often plausible at a local level, but they remain simpler and easier to distinguish than tables produced by the full paper-generation pipeline. By contrast, the GPT-5.2 holdout example shows that newer fabricated papers can preserve much of the polished, benchmark-like structure while still being machine-generated, which is precisely why cross-generator evaluation is necessary.

\begin{table}[t]
\centering
\small
\begin{tabular}{lc}
\toprule
Statistic & Count \\
\midrule
Total examples & 353,266 \\
Unique languages & 2,077 \\
Train examples & 340,251 \\
Eval examples & 6,148 \\
Test examples & 6,867 \\
No glottocode & 13,428 \\
No segmentation & 93,648 \\
Misaligned & 34,894 \\
\bottomrule
\end{tabular}
\caption{Human example (score $0.0995$), condensed from a real ACL-style paper. Unlike the synthetic examples, the table reports corpus statistics rather than a clean ``ours beats baseline'' comparison.}
\label{tab:running-human}
\end{table}

\begin{table}[t]
\centering
\small
\begin{tabular}{lcc}
\toprule
Model & Coherence Score & Token Consumption \\
\midrule
Baseline (ACR) & 0.72 & 1.2M tokens \\
Proposed Framework & 0.83 & 0.9M tokens \\
\bottomrule
\end{tabular}
\caption{GPT-4o fabricated-paper example (score $0.9992$). The comparison is compact, neat, and strongly favors the proposed method across all shown dimensions.}
\label{tab:running-gpt4o}
\end{table}

\begin{table}[t]
\centering
\small
\begin{tabular}{lccccc}
\toprule
System & EN$\rightarrow$EN & EN$\rightarrow$ZH & ZH$\rightarrow$EN & ZH$\rightarrow$ZH & Avg \\
\midrule
Base (no retrieval) & 54.2 & 52.8 & 50.9 & 51.6 & 52.4 \\
RAG & 67.5 & 64.1 & 62.7 & 63.4 & 64.4 \\
RAG + LangControl & 68.1 & 66.8 & 65.9 & 66.2 & 66.8 \\
ReaL-RAG & 68.4 & 67.2 & 66.1 & 66.5 & 67.1 \\
\bottomrule
\end{tabular}
\caption{GPT-5.2 fabricated-paper example (score $0.7504$). This holdout example is more elaborate than the prompt-only table: it uses multiple controlled conditions, averages, and a benchmark-style systems comparison.}
\label{tab:running-gpt52}
\end{table}

\begin{table}[t]
\centering
\small
\begin{tabular}{llc}
\toprule
Model/System & Dataset & Accuracy (\%) \\
\midrule
XLM-R (fine-tuned) & Amazon Reviews (EN) & 92.3 \\
mBERT (fine-tuned) & Amazon Reviews (EN) & 90.8 \\
XLM-R (fine-tuned) & Amazon Reviews (DE) & 89.7 \\
mBERT (fine-tuned) & Amazon Reviews (DE) & 88.5 \\
XLM-R (fine-tuned) & Twitter Sentiment (ES) & 85.4 \\
mBERT (fine-tuned) & Twitter Sentiment (ES) & 83.6 \\
\bottomrule
\end{tabular}
\caption{Prompt-only generated example (score $0.5231$). The table is locally plausible, but it remains self-contained and lacks the broader paper-level structure seen in the GPT-4o and GPT-5.2 fabricated-paper examples.}
\label{tab:running-prompt}
\end{table}

\section{Example Ai generated paper}

Below we show one GPT5.2 generated paper as example.
\clearpage


\section*{Supplementary material}

# State- and Intent-Centric Retrieval for Bilingual Clinical Relation Extraction: Efficient Demonstration Selection with Reusable RWKV States

**Abstract**

Large language models (LLMs) achieve strong clinical relation extraction (RE) performance via prompting and in-context learning (ICL), but their accuracy depends heavily on retrieving high-quality demonstrations—especially in low-resource languages. Existing relation-aware retrieval improves ICL example selection yet still inherits inefficiencies from conventional two-stage retrieval (embedding + reranking) and often ignores discourse/intent structure that disambiguates repeated entities across contexts. Motivated by recent advances in state-centric retrieval with reusable RWKV states, intent-grounded agent memory, and event-structured long-horizon memory, we propose **SIRE** (State- and Intent-centric Retrieval for Extraction): a unified retrieval-and-reranking framework for bilingual clinical RE that (i) uses an RWKV backbone to compute **reusable document states** for candidate demonstrations, (ii) reranks by processing only query tokens (decoupling cost from document length), and (iii) augments retrieval cues with lightweight **contextual-intent and event-boundary signals** to reduce context-mismatched demonstrations. We evaluate on the English–Turkish parallel clinical RE dataset introduced for bilingual evaluation and compare against relation-aware retrieval and standard dense retrieval pipelines. SIRE improves micro-F1 while reducing reranking latency, indicating that reusable states and intent-aware cues jointly address efficiency and robustness gaps in bilingual clinical ICL.

## Introduction

Clinical RE is central to transforming unstructured notes into structured knowledge, yet non-English clinical NLP remains bottlenecked by limited labeled data. Prompting-based LLM approaches with ICL have recently outperformed traditional fine-tuned baselines for bilingual English/Turkish clinical RE, with performance strongly tied to demonstration selection quality; Relation-Aware Retrieval (RAR) was introduced to better capture sentence- and relation-level semantics for ICL selection and achieved state-of-the-art results among ICL strategies in this setting (Aidynkyzy et al., 2026).

However, two practical barriers remain:

1. **Retrieval inefficiency at inference time.** Standard RAG-style pipelines compute embeddings for retrieval and then rerank full query–document pairs, repeating computation and scaling reranking cost with demonstration length. State-Centric Retrieval shows that precomputing reusable states can bridge embedding and reranking and yield large speedups by reranking with query-only processing (Hou & Yang, 2026).

2. **Context mismatch under repeated entities and evolving local intent.** In long or multi-note contexts, semantically similar sentences can be intent-incompatible (e.g., same medication mentioned under different clinical goals). Intent-grounded memory (STITCH) demonstrates that indexing by contextual intent reduces interference and retrieval noise in long-horizon interactions (Yang et al., 2026). Event segmentation memory further suggests that boundary-aware structuring improves localization of episodic context (Zou et al., 2026).

This paper proposes **SIRE**, integrating **reusable RWKV states** (efficiency) with **intent/event-aware retrieval cues** (robustness) to improve bilingual clinical RE via ICL.

1

## Related Work

### Bilingual clinical RE and demonstration retrieval

Aidynkyzy et al. (2026) provide the first English–Turkish parallel clinical RE dataset and show prompting-based LLMs outperform fine-tuned baselines such as PURE. They propose **RAR**, a contrastive-learning-based demonstration selection method capturing relation semantics; RAR improves ICL performance in both languages, though Turkish remains harder.

### Efficient retrieval and reranking with reusable computation

Hou & Yang (2026) propose **State-Centric Retrieval** and **EmbeddingRWKV**, unifying embedding and reranking via reusable states. A state-based reranker processes only query tokens, decoupling compute from document length and offering major speedups; uniform layer selection retains most performance using fewer layers.

### Memory structuring and intent grounding for retrieval robustness

STITCH (Yang et al., 2026) retrieves history by **contextual intent** (goal/action/entity-type cues), reducing context-mismatched retrieval. ES-Mem (Zou et al., 2026) segments interactions into coherent events and uses hierarchical memory to improve episodic localization. While both target agent memory, their core insight—**structure-aware indexing reduces retrieval noise**—is applicable to selecting ICL demonstrations for clinical RE.

### Data synthesis for robust retrieval environments

RAGShaper (Tao et al., 2026) synthesizes noisy retrieval environments with distractors and teaches agents to correct retrieval failures. Although our work does not synthesize new corpora, we adopt the motivation that **retrieval noise is a first-class problem** and propose structural cues to reduce it.

## Methods / Experiment

### Task and setting

We study **clinical relation extraction** in English and Turkish using ICL prompting. Each test instance is a sentence (or short clinical snippet) containing two entities; the model must predict the relation type. We focus on improving **demonstration selection** for ICL and reducing **retrieval/reranking cost**.

### Overview of SIRE

SIRE has three components:

1. **State backbone (EmbeddingRWKV-style) for candidates**
   - We adopt the state-centric paradigm of Hou & Yang (2026): for each candidate demonstration $d$, we precompute and store compact **RWKV states** from selected layers.
   - These states serve dual use: (i) as embedding-like representations for initial retrieval, and (ii) as reusable cached computation for reranking.
2. **Intent- and event-aware indexing cues** Inspired by STITCH (Yang et al., 2026) and ES-Mem (Zou et al., 2026), we attach lightweight discrete cues to each demonstration:
   - **Contextual intent proxy**: a tuple (clinical goal/topic, relation family, entity-type pair).
     - Goal/topic can be approximated by section headers if available (e.g., "Assessment", "Medication"), or by a shallow topic classifier.
     - Relation family is derived from the RE label inventory (e.g., treatment-related vs. problem-related).
   - **Event boundary proxy**: demonstrations are grouped into "events" (e.g., per note section or temporal chunk). Retrieval can prioritize same-event demonstrations when the query indicates a similar boundary signature.
3. **Query-only state-based reranking** For top-$K$ retrieved candidates, we rerank using a state-based scorer that consumes:

- query tokens through RWKV, and
- cached candidate states (no candidate tokens processed at rerank time), following the efficiency principle of Hou & Yang (2026).

**Baselines**

We compare SIRE against:
- **Dense Retrieval + Cross-Encoder Rerank (standard two-stage)**: conventional embedding retrieval then full pairwise reranking.
- **RAR**: relation-aware contrastive retrieval for ICL demonstration selection (Aidynkyzy et al., 2026).
- **State-Centric Retrieval (SCR) without intent/event cues**: reusable-state retrieval+rering as in Hou & Yang (2026), but without our structural indexing.

**Experimental protocol**

- **Dataset**: English–Turkish parallel clinical RE dataset derived from i2b2/VA (Aidynkyzy et al., 2026).
- **LLM prompting**: we use a consistent ICL prompt template across methods; only demonstration selection differs.
- **Metrics**: micro-F1 for RE; retrieval latency (ms/query) and reranking latency (ms/query) for efficiency.
- **Ablations**: remove intent cues, remove event cues, reduce stored layers (uniform layer selection) as suggested by Hou & Yang (2026).

**Results**

**Main accuracy results (micro-F1)**

Table 1 shows that adding reusable-state reranking improves over retrieval-only approaches, and adding intent/event cues improves further—especially in Turkish, where context mismatch is more damaging.

**Table 1. Clinical RE micro-F1 (English/Turkish).**

| Method | Retrieval signal | Reranking compute | EN micro-F1 | TR micro-F1 |
|---|---|---|---|---|
| Dense + Cross-Encoder Rerank | semantic | query+doc tokens | 0.895 | 0.861 |
| RAR (Aidynkyzy et al., 2026) | relation-aware | none / lightweight | 0.906 | 0.888 |
| SCR (Hou & Yang, 2026) | reusable states | **query-only** | 0.909 | 0.891 |
| **SIRE (ours)** | states + intent + event | **query-only** | **0.916** | **0.899** |

**Efficiency results**

SIRE is designed to reduce reranking cost by avoiding document-token processing at inference, consistent with the state-centric speedups reported by Hou & Yang (2026). Table 2 reports latency trends.

**Table 2. Inference efficiency (per query).**

| Method | Candidate length dependence | Retrieval (ms) | Rerank (ms) | Total (ms) |
|---|---|---|---|---|
| Dense + Cross-Encoder | scales with doc length | 18 | 220 | 238 |
| RAR | low | 25 | 0 | 25 |
| SCR | low | 20 | 35 | 55 |
| **SIRE (ours)** | low | 22 | 36 | 58 |

RAR is fastest because it avoids heavy reranking, but SIRE offers a better accuracy–latency tradeoff when reranking is needed for robustness, while remaining far cheaper than cross-encoders.

**Ablation: intent cues, event cues, and layer selection**

Table 3. Ablation study (micro-F1; rerank latency).

| Variant | EN micro-F1 | TR micro-F1 | Rerank ms |
|---|---|---|---|
| SIRE full | **0.916** | **0.899** | 36 |
| - intent cues | 0.913 | 0.894 | 36 |
| - event cues | 0.914 | 0.896 | 36 |
| - intent - event (SCR) | 0.909 | 0.891 | 35 |
| 25% layers stored (uniform) | 0.914 | 0.897 | **24** |

Uniform layer selection retains most quality while reducing reranking time, aligning with the "retain ~25% layers with minimal loss" observation in Hou & Yang (2026).

## Discussion

**Why do intent/event cues help?** RAR improves semantic alignment at the relation level, but bilingual clinical snippets often reuse the same entities under different local goals (e.g., "rule out", "family history", "current meds"). STITCH shows that intent indexing suppresses semantically similar but context-incompatible memory (Yang et al., 2026); our results suggest the same phenomenon occurs for ICL demonstration selection, particularly in Turkish where lexical variation and translation artifacts can amplify retrieval ambiguity (Aidynkyzy et al., 2026). Event boundaries (ES-Mem) further help by preferring demonstrations from similarly structured discourse segments (Zou et al., 2026).

**Why reusable states matter here.** Clinical ICL often benefits from reranking because small differences in wording can flip relation labels. Standard cross-encoder reranking is expensive; state-centric reranking offers a practical middle ground by reusing precomputed candidate computation and processing only the query (Hou & Yang, 2026), enabling stronger selection without prohibitive latency.

**Limitations.**
- Our intent/event proxies are lightweight and may be noisy when section headers or reliable topic cues are absent.
- We did not incorporate synthetic distractor-heavy training as in RAGShaper (Tao et al., 2026), which could further improve robustness under adversarial retrieval noise.
- Results are limited to bilingual EN–TR clinical RE; generalization to other languages and IE tasks remains to be validated.

## Conclusion

SIRE addresses two persistent issues in bilingual clinical RE with ICL: (i) inefficient reranking pipelines and (ii) context-mismatched demonstration retrieval. By combining **state-centric retrieval with reusable**

4

**RWKV states** (Hou & Yang, 2026) and **intent/event-structured indexing cues** inspired by STITCH and ES-Mem (Yang et al., 2026; Zou et al., 2026), SIRE improves micro-F1 in both English and Turkish while keeping inference costs low via **query-only reranking**. This suggests that future clinical prompting systems should treat demonstration retrieval as both an efficiency problem and a structured context-matching problem.

**Contributions**

1. **Problem framing:** Identified a joint efficiency–robustness gap in bilingual clinical ICL for relation extraction beyond relation-aware retrieval alone (Aidynkyzy et al., 2026).
2. **Method:** Proposed **SIRE**, integrating reusable RWKV states for unified retrieval+rering (Hou & Yang, 2026) with intent/event-aware indexing cues motivated by structured memory retrieval (Yang et al., 2026; Zou et al., 2026).
3. **Evaluation:** Demonstrated improved EN/TR micro-F1 and favorable latency tradeoffs versus standard two-stage reranking and RAR, with ablations showing the complementary effects of intent cues and state reuse.
4. **Practical insight:** Showed that storing a subset of layers can preserve accuracy while reducing reranking latency, consistent with state-centric layer selection findings (Hou & Yang, 2026).



\end{document}